\documentclass{article}
\usepackage{ase_preprint,times}
\usepackage{amsmath,amssymb,booktabs,graphicx,tabularx,array}
\usepackage{hyperref,url}
\hypersetup{colorlinks=true,citecolor=blue,linkcolor=blue,urlcolor=blue,
  pdftitle={ASE: Attribution-Guided Semantic Evolution for Time Series Forecasting},
  pdfauthor={Yuxuan Zheng, Zhaoyang Li, Purui Liu, Hengbo Xiao, Guannan He}}

\newcommand{\R}{\mathbb{R}}
\iclrfinalcopy
\title{ASE: Attribution-Guided Semantic Evolution\\for Time Series Forecasting}
\author{Yuxuan Zheng$^{1,\dagger}$ \quad Zhaoyang Li$^{1,\dagger}$ \quad Purui Liu$^{1}$\\
\textbf{Hengbo Xiao}$^{2}$ \quad \textbf{Guannan He}$^{1,*}$\\[0.5em]
$^{1}$Peking University \quad $^{2}$Yoolee.ai\\
$^{\dagger}$Equal contribution. \quad $^{*}$Corresponding author.}
\begin{document}
\maketitle
\begin{abstract}
External text can supply predictive information absent from numerical histories, but its value depends on how it is represented, combined and deployed. We present Attribution-guided Semantic Evolution (ASE), a modular framework for constructing textual factors, evaluating their conditional contribution, and deploying validation-selected numerical--textual predictors. ASE supports task-specific implementations with lexical measurements, semantic primitives or evolved event/state factors, coupled to suitable numerical experts. Its feedback-based implementation uses conditional refitting diagnostics to guide factor revision and combination search. We evaluate ASE on the four text-associated TESS datasets and dynamic CSI 300 stock ranking. The four forecasting implementations obtain test MSEs of 0.001041, 0.022439, 0.333115 and 1.8150, each below the corresponding published TESS score. On CSI 300, the frozen text-fusion implementation improves mean RankIC by 0.000642 over its numerical expert and by 0.000701 over matched shuffled text, with positive differences in all five seeds. Controlled text perturbations support functional reliance on text correspondence, while incumbent retention prevents regressions in the measured development criterion. Together, the experiments characterize ASE at two levels: forecasting performance of task-specific implementations and the conditional contributions of individual modules.
\end{abstract}

\section{Introduction}
Text-augmented forecasting requires more than retrieving relevant documents. A description of rising input costs may be useful when pricing power is weak, redundant when recent prices already reflect the event, or uninformative for the selected prediction horizon. The relevant object is therefore not a universally useful textual feature, but a measurement whose value is conditional on the incumbent predictor and its other inputs. Structured temporal primitives offer one way to bridge text and numerical forecasts \citep{li2026tess}; news-based factor extraction and reflection provide complementary approaches \citep{wang2024llmfactor,wang2024newsforecast}.

We study a different feedback question: \emph{what information should a forecast model return to the language model that proposes its textual measurements?} An aggregate score indicates whether a system improved, but does not identify which definition was redundant, which contribution depended on the current combination, or whether gains persisted across development blocks. Conversely, plausible verbal explanations need not describe what actually helped the predictor. The feedback-based ASE implementation links semantic revision to measured add/drop effects obtained by refitting the same predictor used for forecasting.

In the iterative implementation, the feedback loop alternates two operations. C1 proposes event and state definitions and compiles them into continuous similarity measurements and event--state products. C2 searches numerical--textual combinations under a fixed budget and returns conditional contribution and redundancy diagnostics. The preceding selected combination remains a candidate, preventing search from discarding a useful incumbent simply because a new round restarts elsewhere. The language model edits definitions; forecasting weights are fitted from numerical targets, with no language-model fine-tuning.

\begin{figure}[t]
\centering\includegraphics[width=\linewidth]{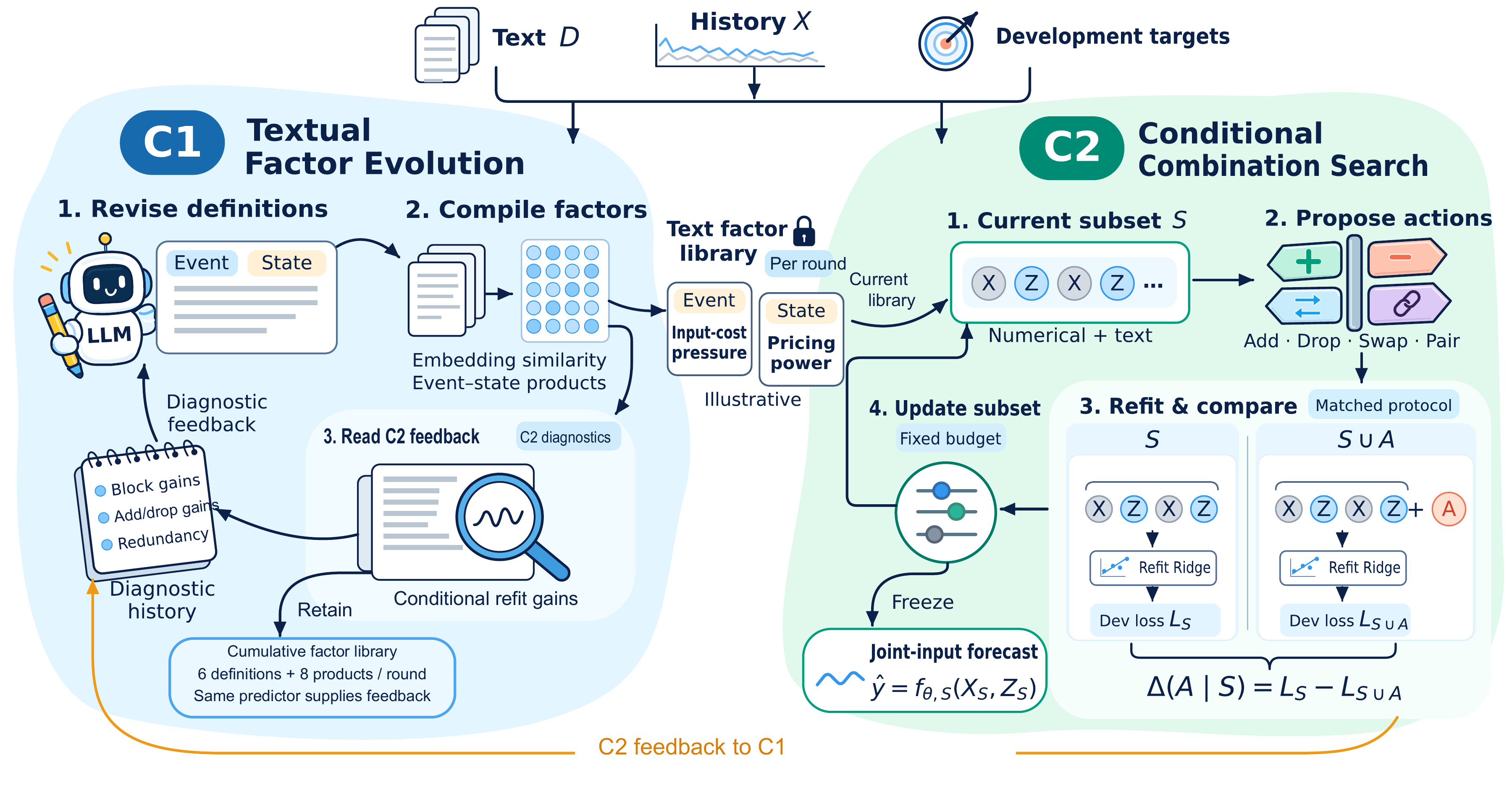}
\caption{The iterative feedback implementation of ASE. C1 proposes and compiles event/state definitions into an accumulated library. C2 searches numerical--textual subsets by refitting the same predictor, then returns conditional contribution and redundancy diagnostics to the next C1 generation. Libraries are fixed within each search round; the final selected predictor is frozen for assessment. Event and state names in the figure are illustrative.}
\label{fig:architecture}
\end{figure}

ASE organizes three modules: C1 constructs and revises textual factors, C2 evaluates their conditional combinations, and C3 deploys a frozen predictor or a validation-trained correction gate. We evaluate four-task forecasting and dynamic stock ranking as system-level applications, and use separate controlled experiments to examine the modules. Task implementations choose the factor representation, numerical expert and deployment rule to suit their data. Figure~\ref{fig:architecture} details the iterative C1--C2 implementation; C3 is described below.

\section{Related Work}
\paragraph{Text and time-series representations.}
Time-LLM reprograms numerical patches for a frozen language model \citep{jin2024timellm}, while TimeCMA aligns numerical and history-derived language representations \citep{liu2025timecma}. External documents can supply additional observations: TimeCAP uses language agents for event prediction \citep{lee2025timecap}, Time-MMD aligns text and numerical data \citep{liu2024timemmd}, LLMFactor extracts stock-relevant factors \citep{wang2024llmfactor}, and From News to Forecast refines news selection through reflection \citep{wang2024newsforecast}. ASE revises the intermediate measurement definitions using conditional predictor feedback.

\paragraph{Structured semantics and forecasting agents.}
TESS maps text to predefined temporal primitives with confidence-aware conditioning \citep{li2026tess}; its released implementation also supplies additive and distillation recipes \citep{tesscode2026}. We distinguish published scores from common-protocol reruns. LAFP searches frozen forecasting-model trajectories using language-model ranking \citep{nguyen2026planner}; AuditForecast updates quantitative base probabilities with situational evidence \citep{cao2026auditforecast}. ASE instead revises a reusable input library from regression-loss diagnostics. These agent systems are related work, not evaluated baselines.

\paragraph{Concept libraries.}
Concept bottleneck models use named intermediate concepts \citep{koh2020cbm}; label-free variants avoid manually labeled concept data \citep{oikarinen2023labelfree}. ESCHER evolves visual concepts using history-sensitive vision-language feedback \citep{sehgal2025escher}. ASE uses conditional loss differences from refitted numerical--textual predictors. Its similarities are inspectable measurements, not verified event labels.

\paragraph{Textual optimization.}
OPRO uses candidate/objective histories \citep{yang2024opro}; Self-Refine and Reflexion use iterative verbal feedback \citep{madaan2023selfrefine,shinn2023reflexion}. ProTeGi combines critiques with beam search \citep{pryzant2023protegi}, TextGrad propagates textual feedback \citep{yuksekgonul2025textgrad}, and GEPA couples reflection with Pareto-based evolution \citep{agrawal2026gepa}. ASE applies measured contribution, block variation and redundancy to forecast-feature definitions; no-feedback, scalar and free-reflection controls test this interface.

\paragraph{Factor discovery and attribution.}
AlphaGen optimizes downstream factor-combination performance \citep{yu2023alphagen}; AlphaForge separates mining from dynamic combination \citep{shi2025alphaforge}. AlphaAgent regularizes originality and hypothesis consistency \citep{tang2025alphaagent}, and FactorMiner uses experience memory \citep{wang2026factorminer}. ASE's conditional refitting belongs to wrapper feature selection \citep{kohavi1997wrappers}; its action-specific loss differences are neither SHAP/SAGE estimates nor causal effects \citep{lundberg2017shap,covert2020sage}.

\paragraph{Evaluation.}
DLinear and PatchTST motivate careful numerical-backbone comparisons \citep{zeng2023linear,nie2023patchtst}. Selection can overfit its own criterion \citep{cawley2010selection}, and text forecasting audits distinguish semantics from structural input effects \citep{sridhar2026semantics}. We separate system-level forecasting scores, matched numerical refits, full-pipeline interventions and frozen text-alignment tests.

\section{Problem and Method}
\label{sec:method}
ASE specifies three interfaces: textual measurement construction (C1), conditional evaluation and combination (C2), and validation-controlled deployment (C3). Implementations choose the measurement family and numerical predictor for their task; iterative revision is used when factor definitions are generated dynamically. We first detail this iterative implementation, which is also the basis of the controlled module studies. The experimental section specifies the lexical, cached-primitive, residual and stock-ranking implementations used in the main benchmarks.

At a prediction origin, history $X$ and available documents $D$ are used to forecast $Y\in\R^H$. A textual definition $p$ specifies an event or state, and a compiler produces $z_p=C(p,D)$. Let $\mathcal P_g$ denote the accumulated, generation-indexed library after round $g$, $\mathcal V_g$ index its compiled features and the numerical inputs, and $S\subseteq\mathcal V_g$ select predictor inputs. Training fits $\theta^*(S)$; a separate development criterion $J$ selects subsets and guides definition revision:
\begin{equation}
\mathcal P_g=\mathcal P_{g-1}\cup G(\mathcal H_{g-1},\mathcal D_{g-1}),\qquad
S_g\approx\arg\min_{S\subseteq\mathcal V_g}J(S;\theta^*(S)).
\end{equation}
$G$ is the language-model proposer, $\mathcal H$ stores prior definition names, and $\mathcal D$ contains measured diagnostics. The approximation denotes budgeted local search. Neither reserved-development nor test outcomes enter $G$ or $J$.

\subsection{Semantic Factor Construction and Evolution (C1)}
In the iterative implementation, Qwen3-4B-Instruct-2507 \citep{qwen2025technical} proposes four event and two state definitions per generation. Each proposal records a name, observable definition, parent and edit. The first generation is shared across all compared arms for a dataset and seed. Later proposals receive prior names, three fixed training-news examples and policy-specific feedback. Neither the generator nor the embedding model is fine-tuned.

Qwen3-Embedding-0.6B \citep{zhang2025embedding} encodes documents and definitions with a 256-token limit. Their normalized embedding dot product is standardized using training statistics and clipped to $[-6,6]$. Six raw factors and eight event--state products are compiled per generation; after five rounds the accumulated library has 70 feature slots. For event $e$ and state $s$, the product is $z_{e,s}=z_e\,\sigma(z_s)$, where $\sigma$ is logistic. This implements a continuous conditioning feature; $\sigma(z_s)$ is not a calibrated probability that a state holds. Invalid outputs occupy zero slots without retries. Embedding similarity is a measurement of relatedness, not a verified assertion that an event occurred. After the library and predictor are frozen, forecasting uses document encoding, similarity compilation and Ridge prediction; definition generation is an offline operation.

\subsection{Conditional Evaluation and Combination Search (C2)}
For development loss $L$, define the contribution of adding $A$ to $S$ as
\begin{equation}
\Delta(A\mid S)=L(\operatorname{fit}(S))-L(\operatorname{fit}(S\cup A)).
\end{equation}
Both predictors are refitted. A positive value identifies a useful addition under the specified predictor and fitting protocol; it is not a causal attribution. For binary subset masks $m,m'\in\{0,1\}^{d_g}$ with $d_g=|\mathcal V_g|$, the interaction controller uses
\begin{equation}
t(m,m')=\frac{(2m-1)^\top(2m'-1)}{d_g},\qquad k(m,m')=t(m,m')+0.5t(m,m')^2.
\end{equation}
A kernel Ridge surrogate with regularization 0.1 on standardized scores predicts measured subset scores to prioritize candidate actions. The main-effect control uses $k=t$, and the random control samples candidates without a surrogate. Thus, ``interaction'' refers to combination-dependent subset utility; the final predictor remains Ridge. 

Each search evaluates the numerical subset first. From round two, it also evaluates the preceding selected subset within the same total budget of 128 evaluations. Controllers share the initialization rule, a 16-evaluation exploration phase, add/drop/swap/pair schedule, pools of 32 proposals and 20\% exploration rate. Their subsequent trajectories can differ once an incumbent or surrogate-guided decision differs. Scores combine mean development-block loss, $0.25$ times its block standard deviation and a $10^{-6}|S|$ subset-size penalty. An accepted combination must improve mean block loss over the numerical reference and at least three of four search blocks. If a newly selected combination fails this guard, the previous guard-eligible incumbent is restored; otherwise the numerical reference is retained. Numerical inputs remain selectable, so overall gains can include numerical feature selection.

\paragraph{Retention property.}
With an unchanged development split and unchanged old feature columns, the returned subset's criterion $J$ cannot increase across rounds. The previous subset is evaluated within the new budget; any accepted search move has lower $J$. If its guard fails, the previous eligible subset is restored. This is an optimization property of the measured criterion, not a guarantee of lower future error. A restart-only control omits incumbent retention.

\subsection{A Predictor-consistent Feedback Loop}
The iterative implementation used for controlled module studies uses a joint Ridge-10 predictor throughout generation feedback, combination search and final prediction. For each history, normalized targets use its mean and standard deviation; training weights proportional to history variance align fitting with original-scale trajectory squared error. Specifically, with history mean $c_i$, standard deviation $s_i$, and $\bar{s^2}=n^{-1}\sum_i s_i^2$, we fit normalized targets with weights $w_i=s_i^2/\bar{s^2}$. For $\hat Y_i=c_i+s_if_\theta(X_i,Z_i)$,
\begin{equation}
\sum_i w_i\left\|\frac{Y_i-c_i}{s_i}-f_\theta(X_i,Z_i)\right\|^2
=\frac{1}{\bar{s^2}}\sum_i\|Y_i-\hat Y_i\|^2.
\end{equation}
Thus, the data-fit term is proportional to original-scale squared error; Ridge regularization is applied to standardized input coefficients with penalty 10. Feature standardization uses training data only. The same predictor therefore supplies conditional diagnostics and final forecasts (Figure~\ref{fig:architecture}).

At each generation, C2 searches the accumulated library. Each factor is then added to or removed from the selected combination and the predictor is refitted. Signed contribution, block variation and training-only redundancy diagnostics, together with the selected combination, form the next C1 feedback. Four high- and four low-scoring diagnostic cards are returned. A positive add gain means adding a measurement reduces error; a positive drop gain means removing it increases error. Neither sign describes the direction of the target. The final revision explicitly states this distinction in every performance-feedback prompt. A retention-only control omits this clarification, separating its effect from incumbent retention. No-feedback generation receives no performance evidence; scalar feedback receives per-factor scores; free reflection receives the same diagnostic evidence without the structured revision instructions.

\subsection{Validation-controlled Deployment (C3)}
C3 freezes the deployed predictor and its validation-selected controls. In residual implementations, it takes the form
\begin{equation}
\hat Y=b(X)+g(X,Z)\sum_k w_k r_k(X,Z),
\end{equation}
where $b$ is the numerical expert, $r_k$ are correction branches and $g$ is either constant or a learned activation rule. The deployment interface accommodates both joint prediction and residual correction. The iterative implementation freezes its selected joint predictor directly; FNSPID uses an always-on lexical correction; Environment uses a gate trained from out-of-fold development gains. The deployment gate controls correction activation, whereas the search-stage guard controls candidate acceptance.

\section{Experimental Setup}
\label{sec:setup}
\paragraph{Four-task forecasting.}
We use FNSPID \citep{dong2024fnspid}, Electricity, Environment and Bitcoin from TESS. Their training/validation/test counts are 6,842/1,424/1,567; 2,955/700/855; 1,108/237/239; and 98/12/13, with forecast horizons 5, 48, 7 and 7. Test MSE is normalized by the global variance of training histories. The main leaderboard combines published TESS/reference scores with locally adapted short-window models, including small PatchTST, iTransformer and TimeMixer implementations. It is a system-level reference comparison, not a claim of identical implementations, tuning budgets or paired predictions across every row. Task-specific model selection uses validation data; the iterative module studies use only its first 75\%, reserving the remainder for assessment. The serialized benchmark rows lack authenticated dates/entity identifiers.

\paragraph{Dynamic CSI 300 ranking.}
Historical index membership covers 627 securities cumulatively, with 299--300 valid observations per trading day. The numerical expert uses Alpha158 and balanced120 features. Training uses 2021, search early 2022, reserved assessment late 2022, and testing 721 label-boundary-purged dates in 2023--2025. Chinese announcement titles map to the next trading day; days without text remain included. Exact intraday timing is not authenticated. RankIC is the daily cross-sectional Spearman correlation between forecasts and realized labels, averaged over test dates. Five paired seeds (2031, 2037, 2053, 2069, 2081) are evaluated, with uncertainty estimated by 2,000 paired 20-day block-bootstrap draws. Training/search text coverage is approximately 20.4\%/20.9\%.

\paragraph{Controlled module experiments.}
The iterative ASE implementation uses Qwen3 proposals, joint Ridge and five final-round seeds (314, 2718, 1618, 5772, 811). Eight feedback/search policies run for five generations with at most 128 search requests per round. Generation uses temperature 0.65, top-$p$ 0.9 and a 1,536-token cap. All jobs freeze before final assessment. Complete-pipeline product removal and mandatory-numerical-input interventions rerun all five rounds. Frozen text-alignment tests use 50 permutations per model seed. Historical test exposure makes these retrospective comparisons, not untouched future-period validation. Complete controls and task-wise outcomes are retained in the research records.

\paragraph{Shared controls and assessment.}
Numerical Ridge and all iterative C2 arms use history-normalized levels, first differences, mean/standard deviation/minimum/maximum of the levels, and mean/standard deviation of the differences. The main-effect controller retains pair actions but removes the quadratic surrogate term. No-feedback runs keep the same generation count and feature-slot capacity. Every arm permits a development-selected numerical fallback. All five seeds are assessed at generation five. Archived paired diagnostics enumerate all $2^5$ sign assignments; they assume exchangeable paired signs and are not multiplicity-adjusted.

\section{Results}
\subsection{Main Forecasting Leaderboard: Four TESS Datasets}
The main leaderboard compares twelve forecasting systems across four datasets. ASE is instantiated for each task using the representations, numerical experts and deployment rules described below. Each column uses the reported target and normalized-MSE convention; baseline sources and protocol differences are specified in the experimental setup.
\begin{table}[htbp]
\centering\small
\setlength{\tabcolsep}{5pt}
\begin{tabular}{lrrrr}
\toprule
Method & FNSPID & Electricity & Environment & Bitcoin \\
\midrule
Persistence & 0.001551 & 0.099490 & 0.518086 & 2.7678 \\
Ridge & 0.001612 & 0.029461 & 0.363842 & 4.0281 \\
NLinear & 0.001512 & 0.028769 & 0.405229 & 2.7600 \\
DLinear & 0.002096 & 0.028570 & 0.361886 & 3.7084 \\
GRU & 0.001615 & 0.029219 & 0.372453 & 4.0124 \\
iTransformer & 0.001725 & 0.025178 & 0.373132 & 3.7583 \\
PatchTST & 0.001705 & 0.024300 & 0.370600 & 4.1126 \\
TimesNet & 0.001631 & 0.024200 & 0.347300 & 3.6724 \\
TimeMixer & 0.001614 & 0.025200 & 0.369300 & 3.9554 \\
Nonstationary & 0.001600 & 0.025600 & 0.347200 & 4.3420 \\
TESS (published) & 0.001200 & 0.023000 & 0.353400 & 2.2726 \\
\midrule
ASE & \textbf{0.001041} & \textbf{0.022439} & \textbf{0.333115} & \textbf{1.8150} \\
\bottomrule
\end{tabular}
\end{table}

The four ASE implementations attain MSEs of \textbf{0.001041, 0.022439, 0.333115 and 1.8150}, respectively, all below the published TESS point estimates. These are system-level point comparisons with published TESS scores; original TESS predictions are unavailable for paired testing.
\begin{figure}[t]
\centering\includegraphics[width=\linewidth]{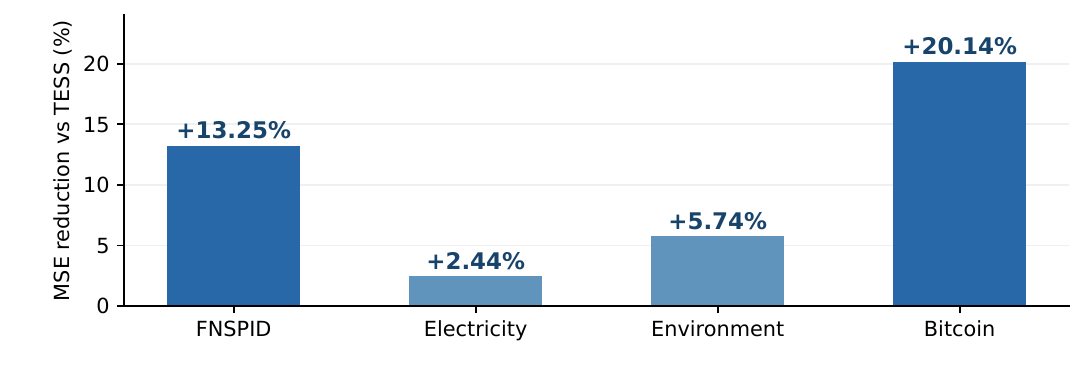}
\caption{MSE reductions of task-specific ASE implementations relative to published TESS scores, computed from the displayed main-table values. The complete twelve-method comparison is retained in the table. These are system-level point comparisons; baseline implementations and selection protocols differ as specified in the setup.}
\label{fig:leaderboard}
\end{figure}

\begin{samepage}
\paragraph{FNSPID: lexical residual correction.}
Numerical Ridge is augmented with lexical, cached-primitive and expanded-factor branches. Validation selects weights $(1.25,0,0)$ and an always-open gate. The resulting MSE falls from 0.001611835 to 0.001041480. The selected ASE configuration therefore deploys the lexical residual branch.
\end{samepage}

\paragraph{Bitcoin: numerical--primitive additive forecasting.}
A 484,243-parameter neural model combines numerical representations with text-derived primitives. Each seed chooses its checkpoint and text scale using validation data. The five test MSEs are 1.855393, 1.838547, 1.739492, 2.131106 and 1.510212; all are below the published TESS value 2.2726. Their mean is 1.814950. This ASE implementation uses cached semantic primitives and validation-selected text scaling; the test set contains 13 windows.

\paragraph{Electricity and Environment: strong numerical experts.}
Electricity uses a nonnegative ensemble of six LightGBM predictors plus a no-intercept Ridge correction from nine selected text/state-interaction factors. Text reduces MSE from 0.022466282 to 0.022438708, a 0.123\% increment. Environment uses trajectory/calendar Ridge with a selectively activated lexical/factor correction, reducing MSE from 0.333236726 to 0.333115111, a 0.036\% increment. The matched numerical references isolate the additional text contribution within each ASE implementation.

\subsection{Main Financial Application: Dynamic CSI 300}
This experiment tests whether aligned text adds ranking information to a fixed numerical expert under historical dynamic membership. Frozen keyword/TF-IDF/SVD text features are fused with the numerical forecast. A matched control shuffles the same text representation while preserving the fusion specification. This stock-specific ASE implementation uses frozen text measurements and validation-selected fusion.
\begin{table}[htbp]
\centering\small
\begin{tabular}{lrrr}
\toprule
Comparison & ASE RankIC & Reference RankIC & ASE gain \\
\midrule
ASE vs numerical expert & 0.054989 & 0.054346 & \textbf{+0.000642} \\
ASE vs matched shuffled text & 0.054989 & 0.054287 & \textbf{+0.000701} \\
\bottomrule
\end{tabular}
\end{table}

\begin{figure}[t]
\centering\includegraphics[width=\linewidth]{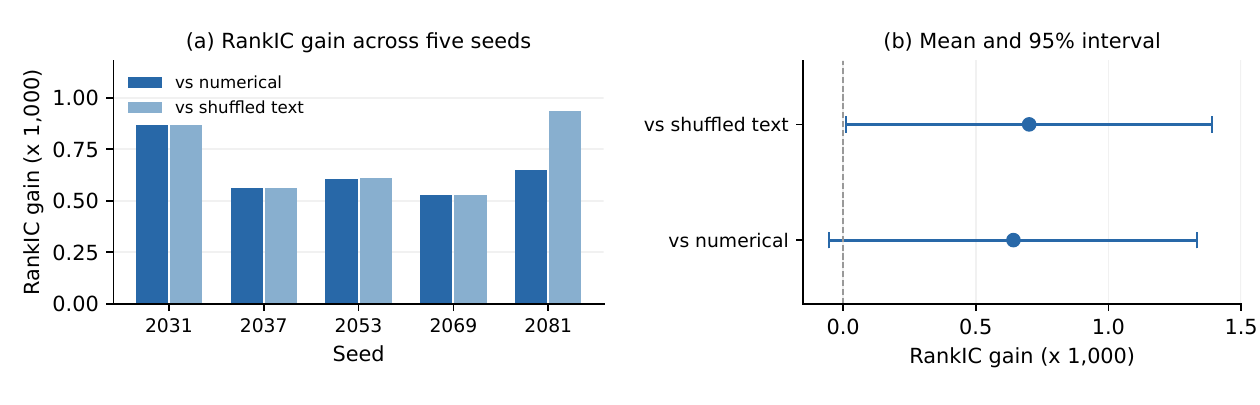}
\caption{CSI 300 text-fusion evidence. Left: paired seed gains over the numerical expert and matched shuffled text. Right: mean gains and paired 20-day block-bootstrap 95\% intervals. The numerical-reference interval includes zero; the matched-shuffle interval is positive under this historical protocol.}
\label{fig:stock}
\end{figure}

Aligned-text fusion improves mean RankIC by \textbf{0.000642} over the numerical expert and by \textbf{0.000701} over matched shuffled text; both comparisons have \textbf{five positive paired seeds}. The respective bootstrap intervals are $[-0.000054,+0.001333]$ and $[+0.000010,+0.001389]$. This separates same-direction seed behavior from temporal uncertainty. The matched-shuffle comparison supports the value of text correspondence within this representation. It is exploratory and not multiplicity-adjusted.

\subsection{Module Analysis}
\paragraph{C1: text correspondence.}
In the iterative ASE implementation, frozen text permutations increase mean MSE by 1.60\%, 1.34\%, 0.43\% and 48.78\% in dataset order. Four of five seed-wise differences are positive in each of the first three tasks, and all five are positive in Bitcoin. This controlled perturbation supports functional reliance on text correspondence; it does not by itself isolate the benefit of factor revision.

\paragraph{C2: retaining useful combinations.}
The restart-only search has 216 increases in the measured development criterion across 640 transitions; both incumbent-retention variants have zero. Retention therefore preserves the best eligible combination under the measured criterion. This is an optimization property, not a guarantee of lower future error.

\paragraph{Task-specific component comparisons.}
On FNSPID, structured revision improves mean MSE over scalar feedback by 2.10\% and free reflection by 2.17\%, with 3/5 paired wins in each comparison. On Electricity, allowing numerical-input selection improves MSE by 2.81\% over retaining all numerical inputs, with 4/5 wins. On Bitcoin, the complete product-bearing pipeline improves MSE by 3.94\% over rerunning without products, with 3/5 wins. These selected task-wise comparisons illustrate conditional benefits: other tasks can favor reduced variants, so they do not establish universal module necessity. Full matched results remain in the research records.

\paragraph{C3: task-specific deployment.}
FNSPID deploys its validation-selected lexical correction on every sample, while Environment activates its correction on 22.6\% of test windows. CSI 300 deploys frozen aligned-text fusion, evaluated above with five paired seeds. These observations characterize deployment choices; a matched gate-removal experiment is not available. Complete task-wise controls are retained in the research records.

\begin{samepage}
\section{Conclusion}
ASE organizes textual forecasting around factor construction, conditional evaluation and controlled deployment. Across four forecasting datasets, task-specific ASE implementations obtain lower MSE point estimates than published TESS scores. The CSI 300 implementation adds positive stock-ranking increments across all five seeds. Controlled module studies characterize the task-dependent roles of structured feedback, conditional selection and incumbent retention. These results support adapting ASE's representation and deployment choices to the forecasting task.
\end{samepage}

\section*{Limitations}
The leaderboard mixes published and locally adapted reference implementations with different backbones and selection protocols. Historical test exposure, unauthenticated chronology and Bitcoin's 13 test windows limit generalization. The stock implementation's numerical-reference interval crosses zero. Module effects vary across tasks; complete controls are retained in the research records, and the gate-ablation gap is stated above.
\section*{Reproducibility Statement}
Sections~\ref{sec:method} and \ref{sec:setup} specify the algorithm, controls and assumptions. The frozen implementation, execution instructions and complete numerical evidence are retained in the research records. Protocols, prompts, source hashes and independent prediction checks are retained in the research archive. Shared calls and conditional evaluation requests are counted explicitly.
\section*{AI Use Statement}
OpenAI Codex assisted method refinement, experimental design and implementation, mathematical exposition, result interpretation, literature retrieval, figure editing and manuscript drafting. The Qwen models specified in Section~\ref{sec:method} generate factor hypotheses and encode text within the evaluated method. AI-assisted code and numerical claims were checked with unit tests, source audits and independent predictor reconstructions. Benchmark targets were not synthesized for this study. The authors take responsibility for the final text, methods, results and references.
\bibliography{references}
\bibliographystyle{iclr2027_conference}
\end{document}